\documentclass[sigconf]{acmart}
\AtBeginDocument{%
  \providecommand\BibTeX{{%
    \normalfont B\kern-0.5em{\scshape i\kern-0.25em b}\kern-0.8em\TeX}}}

\usepackage{soul}

\usepackage{multirow}
\usepackage{enumitem}
\usepackage{microtype}
\usepackage{hyperref}

\usepackage{subcaption}

\newcommand\BibTeX{B\textsc{ib}\TeX}
\newcommand{\stitle}[1]{\noindent{\textbf{#1}}}

\DeclareMathAlphabet{\mathpzc}{OT1}{pzc}{m}{it}

\newcommand{\one} {\mathpzc{1} }
\newcommand{\two} {\mathpzc{2} }

\newcommand{\bigS} {\mathcal{S} }
\newcommand{\bigX} {\mathcal{X} }
\newcommand{\bigY} {\mathcal{Y} }

\newcommand{\smalli} {\mathpzc{i} }
\newcommand{\smallx} {x}
\newcommand{\smally} {y}

\newcommand{\smalln} {n}
\newcommand{\smallj} {j}
\newcommand{\smallk} {k}

\newcommand{\smallr} {r}

\DeclareMathOperator*{\argmin}{arg\,min}
\DeclareMathOperator*{\argmax}{arg\,max}

\newcommand{\ourmodel}{$\mathsf{SCot}$}
\newcommand{\myvalue}[1] {$\mathtt{#1}$}

\newcommand{\myspecial}[1] {\texttt{#1}}

\def\wo{~\mathpzc{w/o}}
\copyrightyear{2023}
\acmYear{2023}
\setcopyright{acmlicensed}\acmConference[WWW '23]{Proceedings of the ACM Web Conference 2023}{May 1--5, 2023}{Austin, TX, USA}
\acmBooktitle{Proceedings of the ACM Web Conference 2023 (WWW '23), May 1--5, 2023, Austin, TX, USA}
\acmPrice{15.00}
\acmDOI{10.1145/3543507.3583541}
\acmISBN{978-1-4503-9416-1/23/04}

\begin{document}

\title{Toward Open-domain Slot Filling via Self-supervised Co-training}


\author{Adib Mosharrof}
\affiliation{%
  \institution{University of Kentucky}
  \city{Lexington}
  \state{KY}
  \country{USA}
}
\email{amo304@g.uky.edu}

\author{Moghis Fereidouni}
\affiliation{%
  \institution{University of Kentucky}
  \city{Lexington}
  \state{KY}
  \country{USA}
}
\email{mfe261@uky.edu}

\author{A.B. Siddique}
\affiliation{%
  \institution{University of Kentucky}
  \city{Lexington}
  \state{KY}
  \country{USA}
}
\email{siddique@cs.uky.edu}

\begin{abstract}
Slot filling is one of the critical tasks in modern conversational systems.
The majority of existing literature employs supervised learning methods, which require labeled training data for each new domain.
Zero-shot learning and weak supervision approaches, among others, have shown promise as alternatives to manual labeling.
Nonetheless, these learning paradigms are significantly inferior to supervised learning approaches in terms of performance.
To minimize this performance gap and demonstrate the possibility of open-domain slot filling, we propose a \textbf{S}elf-supervised \textbf{Co}-\textbf{t}raining framework, called {\ourmodel}, that requires zero in-domain manually labeled  training examples and works in three phases.
Phase one acquires two sets of complementary pseudo labels automatically.
Phase two leverages the power of the pre-trained language model BERT, by adapting it for the slot filling task using these sets of pseudo labels.
In phase three, we introduce a self-supervised co-training mechanism, where both models automatically select high-confidence soft labels to further improve the performance of the other in an iterative fashion. 
Our thorough evaluations show that {\ourmodel}  outperforms state-of-the-art models by 45.57\% and 37.56\% on \myspecial{SGD} and \myspecial{MultiWoZ} datasets, respectively.
Moreover, our proposed framework {\ourmodel} achieves comparable performance when compared to state-of-the-art fully supervised models.

\end{abstract}



\maketitle

\begin{figure}[t!]
\centering
   \includegraphics[width=0.99\linewidth]{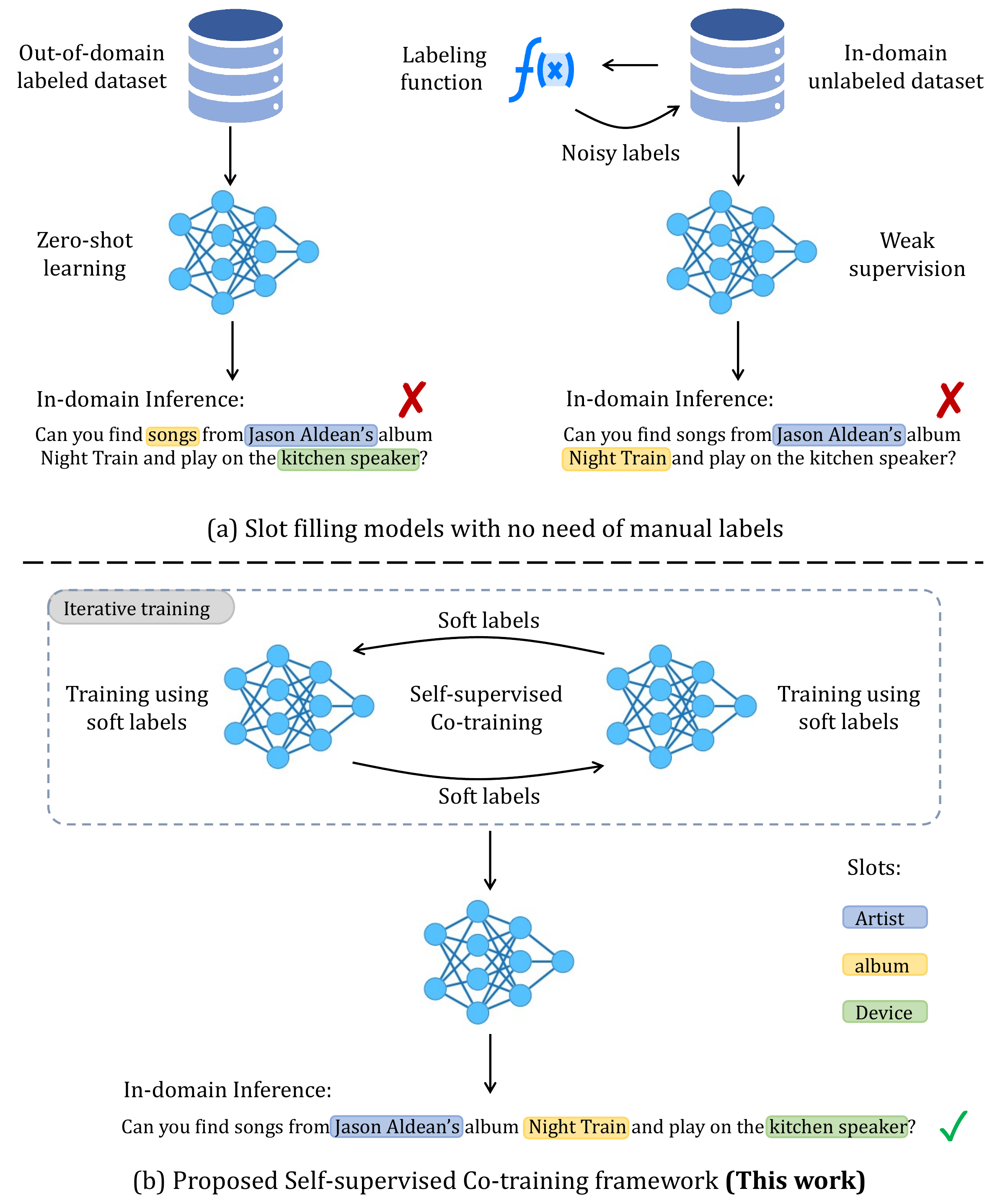}
    \caption{(a) State-of-the-art approaches that do not require manual labels show poor performance as compared to supervised learning models. (b) The proposed self-supervised co-training framework shows comparable performance to state-of-the-art supervised learning models.}
    \label{fig:intro}
    \vspace{-8pt}
\end{figure}

\begin{figure*}[t!]
  \centering
  \includegraphics[width=0.95\linewidth]{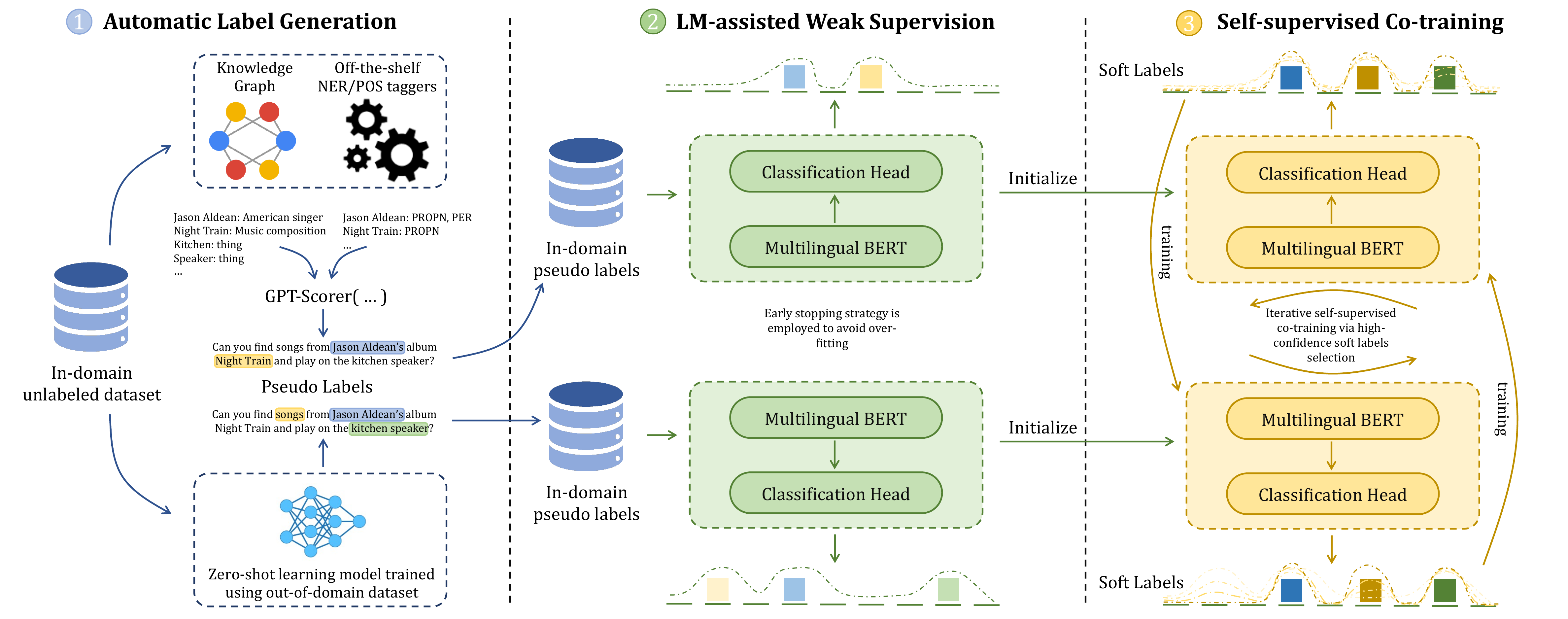}
  \caption{Overview of {\ourmodel}. The self-supervised co-training phase facilitates significant performance gains.  
  }
  \vspace{-12pt}
  \label{fig:intro_model}
\end{figure*}
\section{Introduction}
\label{intro}
Conversational Systems empower users by facilitating interactions through expressive and accessible natural language. 
These systems need to accurately perform \emph{slot filling}, among other tasks, in order to fulfill users’ requests in task-oriented dialog systems or make meaningful recommendations in conversational recommendation systems.
Slot filling is the task of detecting contiguous spans of text that correspond to a slot of interest, usually referred to as a query parameter. 
Figure~\ref{fig:intro} highlights values for various slots.
Due to the vast diversity in users' expressions, slot filling is a challenging task that has been studied extensively in the supervised setting~\cite{goo2018slot,zhang2018joint,young2002talking,bellegarda2014spoken,mesnil2014using,kurata2016leveraging,hakkani2016multi,xu2013convolutional}.
The critical drawback of the supervised learning slot filling methods is that they require a massive amount of labeled training data for each new domain (and slot type) which can be costly and time-consuming. 
This \emph{unscalable} requirement has become a major bottleneck and renders supervised learning methods ineffective because practical conversational systems need to continuously expand their conversational capabilities (e.g., \myspecial{Alexa} \myspecial{Skills}) and be able to converse in new emerging domains.

Recently, researchers have proposed a wide range of approaches to address the label scarcity issue for the slot filling task, such as zero-shot learning~\cite{liu2020coach,siddique2021linguistically} and weak supervision~\cite{hudevcek2021discovering,chen2015jointly}.
Zero-shot learning methods~\cite{siddique2021generalized,siddique2022personalizing} are capable of classifying instances of new unseen classes at inference time, they had not encountered during training, and thus do not require training data for each new domain.
Weak supervision approaches eliminate the need for manually labeled data by automatically generating noisy labels using a heuristic labeling function, powered by almost freely-available external knowledge bases, off-the-shelf models (e.g., NER models), or their combination.
Both learning paradigms have emerged as promising alternatives to manual labeling of the data for every single domain.
However, both approaches suffer from significantly poor performance when compared to the supervised methods.
Motivated by this huge performance gap and the pressing need for slot filling models with no restriction on the domain, we explore the possibility of developing an open-domain slot filling model with comparable performance to the supervised models without incurring the cost and effort associated with manual annotations.

We introduce a \textbf{S}elf-supervised \textbf{Co}-\textbf{t}raining framework, {\ourmodel}, that automatically generates pseudo labels for initial supervision, then it selects high-confidence soft labels to gradually guide the framework to superior prediction performance by leveraging the power of pre-trained language models. 
The proposed framework is also presented in Figure~\ref{fig:intro_model}.
{\ourmodel} employs the pre-trained multilingual BERT and works in three phases.
Phase one acquires two sets of complementary pseudo labels automatically, with no need for human intervention. 
The first set of pseudo labels is generated using a combination of knowledge graphs, off-the-shelf taggers (e.g., NER/POS taggers), and a GPT-2-based~\cite{radford2019language} scorer. 
The other set of pseudo labels is obtained using  a zero-shot slot filling model that has been trained on the out-of-domain dataset and does not require a single in-domain training example. 
Essentially, the first set of labels exploits freely available and domain-independent knowledge sources to acquire labels, whereas the other set of labels leverages the power of zero-shot learning to pick up generalizable patterns of how slot values are mentioned in other domains.

Phase two leverages the power of the pre-trained language models by adapting the multilingual BERT to the task of slot filling.
This phase fine-tunes the model using pseudo labels from phase one with the early stopping strategy to avoid overfitting to the noisy labels.
This phase overcomes the challenge of incomplete and noisy labels to some extent.
Using these sets of pseudo labels from phase one, both models are trained that are capable of making better quality predictions, so pseudo labels from phase one are dropped.
Phase three introduces an iterative self-supervised co-training mechanism that further trains the models on the soft labels.
This phase also proposes a peer training approach where one peer automatically generates soft labels for training the other peer in the next iteration. 
Using this progressive self-supervised training strategy, we can exploit the patterns of slot values in other domains as well as information from external knowledge sources.
Moreover, the high-confidence soft label selection mechanism facilitates efficient learning and enables better model fitting through quality soft labels.
Phase three is the most critical phase that facilitates significant performance gains for the open-domain slot filling task.

Our evaluations using two benchmark datasets, \myspecial{SGD}~\cite{rastogi2019towards} and \myspecial{MultiWoZ}~\cite{zang-etal-2020-multiwoz}, that span up to 20 domains, demonstrate that {\ourmodel} achieves significant performance gains as compared to the learning paradigms~\cite{liu2020coach,siddique2021linguistically,hudevcek2021discovering,chen2015jointly,min2020dialogue,zeng2021automatic} that do not require in-domain labeled data for training. 
Moreover, {\ourmodel} with zero in-domain labeled training examples~\cite{maqbool2022zero}, achieves \emph{competitive} performance as compared to \emph{fully} supervised slot filling models~\cite{chen2019bert,feng2020sequence}.

The contributions of this work are summarized as follows:
\begin{itemize}[leftmargin=1.2\parindent,labelindent=-1pt, itemsep=-1pt]

    \item We propose an iterative self-supervised co-training framework to enable open-domain slot filling that requires zero in-domain manually labeled training examples.

    \item We demonstrate that combining the power of zero-shot learning with external knowledge sources can lead to superior performance and better model fitting.

    \item Our comprehensive experimental analysis using two public datasets \myspecial{SGD} and \myspecial{MultiWoZ} shows that {\ourmodel} outperforms state-of-the-art models consistently and achieves comparable performance to fully supervised models.
\end{itemize}
\section{Preliminaries}
\label{problem}
\subsection{Problem Formulation}

Given a natural language text $\bigX_\smalli = (\smallx_\one, \smallx_\two,\cdots, \smallx_{|\bigX_\smalli|})$, the value for a slot $\bigS_\smallr$ is a contiguous segment of tokens $(\smallx_\smallj,\cdots,\smallx_\smallk)$ such that $0 \leq \smallj \leq \smallk \leq |\bigX_\smalli|$.
Slot filling is generally formulated as a token classification task that predicts the labels $\bigY_\smalli = (\smally_\one, \smally_\two,\cdots, \smally_{|\bigX_\smalli|})$ for the input ${\bigX_\smalli}$ using the \myspecial{IOB} labeling scheme~\cite{ramshaw-marcus-1995-text}. 
Specifically, the first token that points to the value of a slot $\bigS_\smallr$ is provided the label \myvalue{B}-$\bigS_\smallr$, the other tokens inside the slot value are given the label \myvalue{I}-$\bigS_\smallr$, whereas non-slot tokens are labeled as \myvalue{O}.
In the traditional supervised setting, a dataset with $\smalln$ token-level labeled training examples $\{(\bigX_\smalli, \bigY_\smalli)\}_{\smalli=1}^\smalln$ is provided to train the model, whereas acquiring such dataset for each new slot is expensive and labor-intensive. In this work, we focus on a challenging problem setup where we are  provided with unlabeled training examples $\{\bigX_\smalli\}_{\smalli=1}^\smalln$ only, and corresponding labels $\{\bigY_\smalli\}_{\smalli=1}^\smalln$ are not available.
In contrast to the named entity recognition (NER) task where only a limited set of entities (e.g., \myspecial{LOC}, \myspecial{PER}, \myspecial{ORG}) is considered, the slot filling task is more demanding because a wide range of slot types (e.g., \myspecial{price}, \myspecial{destination}, \myspecial{departure}\_\myspecial{time} in the \myspecial{Buses} domain) may appear in the natural language text.

\subsection{Pre-trained NLP Models}
\label{sec:nlpmodels}
This work uses several pre-trained and readily-available natural language processing (NLP) models.
In the following, we provide a brief overview of these models.

\stitle{Pre-trained POS tagger.}
The POS tagger is trained to predict part-of-speech tags for each token in a given natural language text, such as \myspecial{PROPN}, \myspecial{VERB}, and \myspecial{ADJ}.
The POS tags provide useful syntactic cues for slot value identification from natural language texts.
Since POS tagger is domain-independent, it facilitates automatic label generation for the task of open-domain slot filling.
For example, the values for slots are generally tagged as proper nouns by the POS tagger. 
We use the production-ready pre-trained POS tagger\footnote{\url{https://spacy.io/api/annotation\#pos-tagging}} from SpaCy that has produced a high performance for the task.

\stitle{Pre-trained NER model.}
The NER model tags each token in the natural language text for the mentions of the four entities: \myspecial{LOC}, \myspecial{PER}, \myspecial{ORG}, and \myspecial{MISC}.
Although a limited number of entities are tagged by the NER model, these entities can be used to provide cues for many slots (though not all).
For example, \myspecial{LOC} can be used to narrow down the search to \myspecial{source} or \myspecial{destination} slots in the \myspecial{Buses} domain.
Nevertheless, it still proves challenging to automatically assign labels to the given natural language text for the slot filling task.
Furthermore, open-domain NER models are still evolving and may miss mentions of the entities, especially slots of interest in the target domain, which results in a low recall. 
Nonetheless, the NER model facilitates automatic labeling and reduces the task's complexity.
We leverage the pre-trained NER model\footnote{\url{https://spacy.io/api/annotation\#named-entities}} from SpaCy.

\stitle{Pre-trained GPT-2.} 
GPT-2~\cite{radford2019language} is trained for predicting the next token, i.e., autoregressive objective, using 40 GB textual data of WebText dataset in an unsupervised setting.
It also follows transformer architecture~\cite{vaswani2017attention}. 
For a given natural language sequence $(\smallx_1, \cdots , \smallx_{|\smallx|})$ where token $\smallx_\smalli$ is sampled from a fixed vocabulary of tokens, factorizing the joint probabilities over tokens as a product of conditional probabilities~\cite{bengio2003neural} leads to $ \prod_{i=1}^{|\smallx|} p(\smallx_\smalli | \smallx_1, \cdots ,\smallx_{\smalli-1})
$.
This approach not only enables tractable sampling, but also allows for computing $p(\smallx)$, and estimating any conditionals of the form: $p(\smallx_{\smalli - k}, \cdots , \smallx_\smalli | \smallx_1, \cdots , \smallx_{\smalli - k - 1} )$.
In this work, we use the pre-trained GPT-2 model\footnote{\url{https://huggingface.co/gpt2}} to score a given natural language sequence by computing the log probabilities of the tokens to select the best hypothesis for automatic label generation in phase one of {\ourmodel}.

\stitle{Pre-trained BERT.}
BERT~\cite{DBLP:journals/corr/abs-1810-04805} is an unsupervised model that follows the transformer architecture~\cite{vaswani2017attention} by conditioning on bidirectional contexts. 
In contrast to traditional unidirectional language modeling~\cite{bengio2003neural}, BERT has been trained to condition on both left and right contexts simultaneously across all layers to learn deep bidirectional representations from the unlabeled text.
It is trained for masked language model (MLM) and the next sentence prediction (NSP) task. 
In the MLM task, for a given natural language sequence $(\smallx_1, \smallx_2, \cdots, [MASK], \cdots, \smallx_{|\smallx|})$, the goal is to predict $[MASK]$ tokens successfully, whereas the next sentence prediction task focuses on whether sentence B naturally follows sentence A, given two natural language sequences $(\smallx_1, \cdots, \smallx_{|\smallx|})$ and $(\smallx^\prime_1, \cdots, \smallx^\prime_{|\smallx^\prime|})$, using the BooksCorpus (i.e., 800 million words)~\cite{zhu2015aligning} and textual data from Wikipedia English articles (i.e., 2,500 million words). 
The pre-trained BERT has produced state-of-the-art results on many natural language understanding benchmarks, including slot filling task ~\cite{chen2019bert} in the supervised setting.
In our proposed framework {\ourmodel}, we employ the pre-trained multilingual BERT\footnote{\url{https://huggingface.co/bert-base-multilingual-cased}} as the backbone model and adapt it for open-domain slot filling task.

\subsection{Knowledge Graphs}
Knowledge graphs (KG) are used to store information about entities or concepts and are often used as knowledge sources for a range of NLP tasks. 
In this work, we use Google Knowledge Graph Search API
\footnote{\url{https://developers.google.com/knowledge-graph}} to retrieve type information about entity mentions in the natural language.
Information from KG facilitates the generation of several hypotheses about slot labels in phase one of the framework.
Moreover, matching the n-grams in the natural language text to the KG entities also assists in improving recall for the cases where entities or concepts are missed by POS and NER taggers.

\subsection{Zero shot Learning}
Zero-shot learning models have the capability to generalize to new unseen classes, not encountered during training.
Zero-shot learning is one of the promising approaches to address the data scarcity issue.
A few zero-shot learning models~\cite{liu2020coach,shah2019robust,bapna2017towards,siddique2021linguistically} have been proposed for the slot filling task in the literature that has shown encouraging results.
Yet, there exists a huge performance gap compared to the supervised slot filling models.
Nonetheless, zero-shot slot filling models can generate decent quality labels (though noisy) for new unseen domains and slots at \emph{zero cost}.
In this work, we pre-train a zero-shot slot filling model~\cite{liu2020coach} on the out-of-domain slot filling dataset. 
Since the zero-shot slot filling model can infer labels for unseen classes, we use the trained model to generate pseudo labels for the target domain/dataset in phase one.

\section{Proposed Framework: {\ourmodel}}
\label{model}

In this section, we introduce a novel self-supervised co-training framework {\ourmodel}.
The proposed framework is illustrated in Figure~\ref{fig:intro_model} and works in three phases.
Phase one automatically generates two sets of pseudo labels.
We use a combination of off-the-shelf pre-trained POS and NER taggers, knowledge graph, and GPT-2 scorer for generating the first set of pseudo labels automatically without any hand-crafted rules for matching the slot values.
The other set of pseudo labels is acquired through a zero-shot slot filling model~\cite{liu2020coach}, trained on the out-of-domain dataset.
It is critical to emphasize that both sets of labels are noisy and incomplete which poses serious challenges to training effective models for the task of open-domain slot filling.
Phase two fine-tunes the pre-trained BERT to the slot filling task that effectively transfers the knowledge from the pre-trained language model~(LM) to overcome the issue of label incompleteness to some extent. 
Further, we employ the early stopping technique to minimize the noise in the labels.
The output of this phase is two BERT models that can generate soft labels for self-supervision during co-training in phase three.
Phase three leverages the fine-tuned models and further trains them in an iterative fashion.
Specifically, the proposed peer training approach facilitates high-confidence soft label selection for the other peer to perform training. This phase progressively reduces the noise in the labels and enables effective model fitting.

\subsection{Phase One: Automatic Label Generation}
To acquire the first set of labels, we perform the following steps.
First of all, off-the-shelf trained POS and NER taggers are used to predict initial estimates of the slot values irrespective of the slot types. Then, the type information of the slot values is queried from the KG and the slot value is tagged for the most appropriate slot in the target domain.
This approach, however, produces low recall. 
To expand the candidate slot values, we generate n-grams of the natural language text and employ a partial matching scheme to query the KG for type information (e.g., \myspecial{Jason} \myspecial{Aldean} = \myspecial{American} \myspecial{singer}) of the n-grams if the entry exists.
This process generates multiple overlapping hypotheses about the slot values.
We replace a span of text that corresponds to a slot value by its type information and a GPT-2 based scorer (see Section~\ref{sec:nlpmodels}) is used to select the best candidate based on the fluency of the text.
Naturally, if a token (or span of tokens) is replaced by its type, the sentence should score higher as compared to the case where an inappropriate substitution is performed. 
We select the best hypothesis if the score is greater than the threshold.
Intuitively, the candidate selection threshold can automatically be searched based on a small validation set from the target domain, making the label generation process fully automatic. 
The other set of noisy labels is acquired by the zero-shot slot filling model~\cite{liu2020coach} that has been trained using an out-of-domain dataset. It is important to highlight that the zero-shot slot filling model does not require any labeled in-domain training example. 
To summarize the automatic label generation phase, both sets of labels are acquired in a fully automatic fashion without any hand-crafting.

In contrast to previous work in weak supervision~\cite{ren2015clustype,he2017autoentity,fries2017swellshark,giannakopoulos2017unsupervised} that obtains a single set of noisy labels and then propose techniques to overcome the challenge of fitting an effective model to the noisy labels, we acquire two sets of complementary labels.
The choice of these two sets of labels is guided by the intuition that they should be complementary and the models trained on these sets of labels should be able to share complementary information with the other to improve the performance in the later phases of the framework.
Essentially, the first set of labels carries information from external knowledge sources, whereas the labels generated through the pre-trained zero-shot slot filling model capture how the slot values are mentioned in other domains.
To further elaborate on the motivation and our process for the first set of labels (i.e., labels using KG and other NLP models), the pre-trained LMs have been shown to have a great deal of knowledge~\cite{petroni2019language}, thus should be capable of generating automatic labels with no need of external KG. 
To the best of our knowledge, there exists no work that shows that accurate token-level automatic labeling (e.g., slot filling task) is possible with pre-trained LMs. 
Moreover, such approaches would require heavy prompting in each new target domain, whereas our label generation process is fully automatic and only relies on the readily-available pre-trained NLP models and external KG.

\subsection{Phase Two: LM-assisted Weak Supervision}
Since we do not have access to dataset $\{(\mathbf{X}_n,\mathbf{Y}_n)\}_{n=1}^N$ with true ground-truth labels.
We use pseudo labels generated in phase one, $\{(\mathbf{X}_n,\mathbf{D}_n)\}_{n=1}^N$, to learn 
$f_{m,c}(\cdot; \cdot)$ that outputs the probability of the $m$-th token to take on class $c$. 
We learn $f_{m,c}(\cdot; \cdot)$ by minimizing the following loss over the noisy dataset $\{(\mathbf{X}_n,\mathbf{D}_n)\}_{n=1}^N$: 
$$
\hat\theta = \argmin_{\theta}\frac{1}{N}\sum_{n=1}^{N} \ell(\mathbf{D}_n, f(\mathbf{X}_{n}; \theta)),
\label{eq:stage1}
$$
where $\ell(\mathbf{D}_n, f(\mathbf{X}_{n}; \theta)) = \frac{1}{M} \sum_{m=1}^{M} -\log{f_{m,d_{n, m}}(\mathbf{X}_{n}; \theta)}$. 
We employ the pre-trained multilingual BERT with token-level classification head that uses Adam optimizer \cite{kingma2014adam,Liu2019} with early stopping and multiple random initializations.

Since slot filling task is similar to the MLM training objective of the BERT, we employ pre-trained BERT as the backbone model.
That is, MLM's goal is to predict the masked tokens using bidirectional contexts. Similarly, slot filling tries to predict the label for a token leveraging both left and right contexts simultaneously, which makes the pre-trained BERT an ideal model of choice that greatly facilitates minimizing incomplete labels.
It is important to highlight that our automatically generated labels are not only incomplete but also potentially wrong.
The training strategies employed in this phase minimize the noise in the label to some extent. 
Specifically, early stopping can provide a strong regularization and would not let the model overfit to the noisy labels, especially wrong labels. 
Moreover, early stopping does not let the model forget the knowledge in the pre-trained model.
Similarly, multiple random initializations enforce robustness. 
Since the model is fine-tuned on the noisy labels, averaging the predictions of multiple models for each token ensures that wrong labels end up with low probabilities and true labels consistently achieve high probabilities.
Using the above-mentioned strategies, we train two slot filling models, which we call the peers. The peer one is trained on the first set of pseudo labels that were generated using POS and NER taggers, KG, and the GPT-2 scorer in phase one. Similarly, peer two is trained using the predictions of the zero-shot slot filling model~\cite{liu2020coach}.
Both models have the same architecture and follow the same training procedures.

\begin{table*}[t!]
\centering
\caption{Dataset statistics.}
\vspace{-7pt}
\label{tab:dataset}
\begin{tabular}{lccccc}
\toprule
\textbf{Dataset}  & \textbf{Dataset Size} & \textbf{Vocab. Size} & \textbf{Avg. Length} & \textbf{\# of Domains} & \textbf{\# of Slots} \\ \hline
\textbf{SGD}      & 188K                  & 33.6K                & 13.8                 & 20                     & 240                  \\
\textbf{MultiWoZ} & 67.4K                 & 10.5K                & 13.3                 & 8                      & 61 \\
\bottomrule
\end{tabular}
\vspace{-7pt}
\end{table*}

\subsection{Phase Three: Self-supervised Co-training}
We introduce an iterative peer training algorithm where both peers generate high-confidence soft labels for training the other peer in the next iteration. 
Theoretically, these peers can be anything, but in this work, 
we explore two of the most promising directions that have shown the promise to minimize the need for manual labeling for the task: zero-shot learning and distant supervision.
This phase uses a self-supervised co-training scheme to exploit the patterns of slot values from other domains through the labels generated by the zero-shot filling model (i.e., peer two)~\cite{liu2020coach} as well as utilize the knowledge in external KGs and pre-trained models via labels provided by the peer one.
Specifically, we initialize the peers trained in phase two and use their pseudo labels to kick-start training in this phase.
Specifically, peer one $f_{m,c}(\cdot; \theta_{\textrm{p1}})$ would generate labels $\{\tilde{\mathbf{Y}}^{(t)}_n = [\tilde{y}_{n,1}^{(t)}, ..., \tilde{y}_{n,m}^{(t)}]\}_{n=1}^{N}$ for peer two $f_{m,c}(\cdot; \theta_{\textrm{p2}})$ at the $t$-th iteration by:
$$
\tilde{y}_{n,m}^{(t)} = \argmax_{c}{f_{m,c}(\mathbf{X}_n; \theta_{\textrm{p1}}^{(t)})}. 
\label{eq:pseudo}
$$

Based on these labels, the peer two can be fine-tuned by: 
$$
\hat\theta_{\textrm{p2}}^{(t+1)} = \argmin_{\theta}\frac{1}{N}\sum_{n=1}^N \ell(\tilde{\mathbf{Y}}_n^{(t)}, f(\mathbf{X}_{n}; \theta)).
\label{eq:self_train1}
$$

Similarly, peer two $f_{m,c}(\cdot; \theta_{\textrm{p2}})$ would generate pseudo labels for peer one $f_{m,c}(\cdot; \theta_{\textrm{p1}})$ that are used to fine-tune peer one. 
We also notice that it is beneficial to stop early during this phase as well, to improve the model fitting and gradually reduce the noise associated with the automatically generated labels.
Since pseudo labels are refined gradually in an iterative way, both peers can benefit from the knowledge contained within the labels of the other while avoiding overfitting.
Furthermore, as an alternative to pseudo labels, we also generate soft labels that are used for confidence re-weighting. 
The high-confidence soft label selection strategy enables better model fitting and efficient learning via better quality of the automatic labels.
Specifically, for the given $m$-th token in the $n$-th training example, the probability for all classes $C$ is $[f_{m,1}(\mathbf{X}_n;\theta),...,f_{m,C}(\mathbf{X}_n;\theta)]$. 
Following ~\cite{xie2016unsupervised}, at $t$-th iteration, peer one generates soft labels, $\{\mathbf{S}_n^{(t)} = [\mathbf{s}_{n,m}^{(t)}]_{m=1}^M \}_{n=1}^N$, as given below:
$$
\mathbf{s}_{n,m}^{(t)} = [s_{n,m,c}^{(t)}]_{c=1}^{C} = \Bigg[  \frac{f_{m,c}^2(\mathbf{X}_n;\theta_{\textrm{peer1}}^{(t)})/p_{c}}{\sum_{c'=1}^C f_{m,c'}^2(\mathbf{X}_n;\theta_{\textrm{peer1}}^{(t)})/p_{c'}}\Bigg]_{c=1}^{C}
\label{eq:soft}
$$ 
where $p_{c} = \sum_{n=1}^N \sum_{m=1}^M f_{m,c}(\mathbf{X}_n;\theta_{\textrm{p1}}^{(t)})$ computes the frequency of the tokens for the $c$-th class. 
Then, peer two $f(\cdot; \theta_{\textrm{p2}}^{(t+1)})$ is fine-tuned by:
$$
\theta_{\textrm{p2}}^{(t+1)} = \argmin_{\theta} \frac{1}{N} \sum_{n=1}^{N} \ell_{\rm KL}(\mathbf{S}_n^{(t)}, f(\mathbf{X}_{n}; \theta)),
$$
where $\ell_{\rm KL}(\cdot,\cdot)$ is the KL-divergence-based loss:
$$
\ell_{\rm KL}(\mathbf{S}_n^{(t)}, f(\mathbf{X}_{n}; \theta))=\frac{1}{M}\sum_{m=1}^M\sum_{c=1}^C - s_{n,m,c}^{(t)} \log f_{m,c}(\mathbf{X}_{n}; \theta).
\label{eq:klloss}
$$

Moreover, we also investigate selecting tokens that have high confidence. 
For instance, we pick high-confidence tokens from the $m$-th input example at the $t$-th iteration by  
$
H^{(t)}_n = \{m : \max_{c} s_{n,m,c}^{(t)} > \epsilon \},
$
where $\epsilon\in [0,1]$ is a threshold that can be searched based on a small validation set. 
Then, peer two $f(\cdot; \theta_{\textrm{p2}}^{(t+1)})$ is fine-tuned by:
$$
\theta_{\textrm{p2}}^{(t+1)} 
= \argmin_{\theta} \frac{1}{N|H^{(t)}_n|}\sum_{n=1}^{N} \sum_{m\in H^{(t)}_n}\sum_{c=1}^C - s_{n,m,c}^{(t)} \log f_{m,c}(\mathbf{X}_{n}; \theta).
$$

This phase improves the robustness to effectively fit the model for tokens with high confidence. 
Both peers keep sharing information and their confidence by producing soft labels for their counterparts until they approximate to the true labels while employing early stopping and scheduled learning rates.
It is important to remind that phase three is the most important phase that progressively reduces noise from the labels to a great extent and enables superior performance for the task of open-domain slot filling.
\section{Experimental Setup}
\label{experiments}
\subsection{Datasets}
\label{datasets}
We conducted extensive experiments to evaluate the performance of our model {\ourmodel} using two well-known public benchmark datasets: Schema Guided Dialogue~(\myspecial{SGD})~\cite{rastogi2019towards} and Multi-Domain Wizard-of-Oz~(\myspecial{MultiWoZ})~\cite{zang-etal-2020-multiwoz}.
The important statistics about datasets are presented in Table~\ref{tab:dataset}.

\stitle{SGD~\cite{rastogi2019towards}:}
\myspecial{SGD} is one of the most challenging and comprehensive publicly available datasets for evaluating tasks related to dialog systems. 
It consists of dialog utterances from 20 domains and covers 240 slots. Following ~\cite{siddique2021linguistically}, we pre-processed the dataset for the task of slot filling. 
Moreover, all the domains that have less than 1850 utterances (i.e., $<~1\%$ of the dataset) have been merged into a single domain, named ``\myspecial{Others}'' in our experiments.

\stitle{MultiWoZ~\cite{zang-etal-2020-multiwoz}:}
\myspecial{MultiWoZ} has been extensively used to evaluate a wide range of tasks in dialog systems.
In our experiments, we used \myspecial{MultiWoZ} 2.2 which consists of dialog utterances for 61 slots in 8 domains. 
We followed the experimental setup of ~\cite{siddique2021linguistically} and merged all the utterances into a single domain, called ``\myspecial{Others}'', which accounted for $<~1\%$ of the dataset.

\begin{table*}[t!]
\footnotesize
\centering
\caption{Slot F1 score for all the domains in SGD dataset. The zero-label proposed framework {\ourmodel} achieves competitive performance to the fully supervised methods for each domain in the dataset.}
\vspace{-7pt}
\label{tab:sgd-domains}
\begin{tabular}{l|cc|cc|cc|c|cc|c}
\toprule
\multicolumn{1}{c|}{}        & \multicolumn{2}{c|}{\textbf{Unsupervised Learning}}                  & \multicolumn{2}{c|}{\textbf{Weak Supervision}}                             & \multicolumn{2}{c|}{\textbf{Zero-shot Learning}}                & \multicolumn{1}{c|}{\textbf{Few-shot}} & \multicolumn{2}{c|}{\textbf{Supervised Learning}}                        & \multicolumn{1}{c}{\textbf{Self-supervised}} \\
\multicolumn{1}{c|}{\textbf{Domains $\downarrow$}} & \multicolumn{1}{c}{\textbf{DSI}} & \multicolumn{1}{c|}{\textbf{RCAP}} & \multicolumn{1}{c}{\textbf{Inter-Slot}} & \multicolumn{1}{c|}{\textbf{Merge-Select}} & \multicolumn{1}{c}{\textbf{Coach}} & \multicolumn{1}{c|}{\textbf{LEONA}} & \multicolumn{1}{c|}{\textbf{GPT-3}}    & \multicolumn{1}{c}{\textbf{JointBERT}} & \multicolumn{1}{c|}{\textbf{Seq2Seq-DU}} & \multicolumn{1}{c}{\textbf{{\ourmodel} (this work)}}            \\ \hline
Buses                        & 0.2533                  & 0.3705                            & 0.3006	& 0.2860                             & 0.6281                    & 0.6978                     & 0.5054                        & 0.9161                        & \textbf{0.9318}                          & \underline{0.7701}                              \\
Calendar                     & 0.3963                  & 0.5253                            & 0.4810	&0.5134                            & 0.6023                    & 0.7436                     & 0.6269                        & \textbf{0.9862}                        & 0.9605                          & \underline{0.9725}                              \\
Events                       & 0.2778                  & 0.5162                            & 0.2562	&0.3922                            & 0.5486                    & 0.7619                     & 0.6658                        & \textbf{0.9403}                        & 0.9346                          & \underline{0.8773}                              \\
Flights                      & 0.3605                  & 0.4154                            & 0.4240	&0.5220                             & 0.4898                    & 0.5901                     & 0.7166                        & \textbf{0.9612}                        & 0.9286                          & \underline{0.9026}                              \\
Homes                        & 0.4356                  & 0.4747                            & 0.3558&	0.5586                            & 0.6235                    & 0.7698                     & 0.7734                        & 0.9498                        & 0.9860                           & \underline{\textbf{0.9862}}                              \\
Hotels                       & 0.2630                   & 0.3584                            & 0.1566 &	0.4252                            & 0.7216                    & 0.7677                     & 0.6825                        & \textbf{0.9219}                        & 0.9096                          & \underline{0.8542}                              \\
Movies                       & 0.3407                  & 0.3631                            & 0.1414&	0.5420                             & 0.5537                    & 0.7285                     & 0.5377                        & \textbf{0.9423}                        & 0.9281                          & \underline{0.8728}                              \\
Music                        & 0.4146                  & 0.5613                            & 0.3134&	0.4844                            & 0.5786                    & 0.7613                     & 0.6098                        & 0.9387                        & 0.9373                          & \underline{\textbf{0.9419}}                              \\
RentalCars                   & 0.3661                  & 0.4964                            & 0.1664&	0.4114                            & 0.6576                    & 0.7389                     & 0.5339                        & 0.8753                        & \textbf{0.9232}                          & \underline{0.9123}                              \\
Restaurants                  & 0.3537                  & 0.4457                            & 0.1316&	0.6534                            & 0.7195                    & 0.7574                     & 0.5513                        & \textbf{0.9534}                        & 0.9345                          & \underline{0.8928}                              \\
RideSharing                  & 0.3642                  & 0.5164                            & 0.2860	&0.6628                            & 0.7273                    & 0.8172                     & 0.7589                        & 0.9490                         & 0.9429                          & \underline{\textbf{0.9712}}                              \\
Services                     & 0.3902                  & 0.5834                            & 0.3342&	0.6598                            & 0.7607                    & 0.8182                     & 0.5819                        & \textbf{0.9714}                        & 0.9151                          & \underline{0.9688}                              \\
Travel                       & 0.3606                  & 0.5894                            & 0.1950&	0.4618                            & 0.8403                    & 0.9234                     & 0.6715                        & 0.9344                        & 0.9226                          & \underline{\textbf{0.9552}}                              \\
Weather                      & 0.3498                  & 0.5363                            & 0.1484&	0.6390                            & 0.6003                    & 0.8223                     & 0.7040                         & 0.9820                         & 0.8981                          & \underline{\textbf{0.9821}}                              \\
Others                       & 0.3305                  & 0.4323                            & 0.2824&	0.5026                            & 0.4921                    & 0.5592                     & 0.6660                         & \textbf{0.9662}                        & 0.9353                          & \underline{0.9291}                              \\ \hline
Average                      & 0.3519                  & 0.4823                            & 0.2636&	0.5151                            & 0.6466                    & 0.7642                     & 0.6371                        & \textbf{0.9444}                        & 0.9324                          & \underline{0.9186}  \\
\bottomrule
\end{tabular}
\vspace{-7pt}
\end{table*}

\subsection{Evaluation Methodology}
\label{testing}

The Slot F1 score is the standard metric for evaluating the slot filling task, that has been used in our experiments.
We present our evaluations for the following settings.

\stitle{Evaluation using a single domain in the dataset.}
We evaluate using each domain separately in both datasets.
The zero-shot learning model is trained using all the domains in the dataset except the target domain (i.e., being used for evaluation) and the model's predictions for the target domain are used as one set of automatic labels.
Since there exists a wide range of domains in both datasets, the evaluation is used to validate the robustness of the proposed framework across domains.

\stitle{Evaluation using the full dataset.}
We also perform evaluations using full datasets. 
In this setup, the zero-shot learning model is trained using the dataset that is not being used for evaluation.
For example, when we conduct an evaluation on \myspecial{SGD} dataset, the zero-shot slot filling model is trained on \myspecial{MultiWoZ}, and its predictions for the \myspecial{SGD} dataset are considered as automatic labels.

\subsection{Competing Methods}
\label{baselines}
We compare {\ourmodel} with state-of-the-art (SOTA) methods that employ a variety of learning paradigms, including supervised learning for the slot filling task.

\begin{description}[leftmargin=1.2\parindent,labelindent=-3.5pt, itemsep=-1pt]
\item \textbf{DSI~\cite{min2020dialogue}:}
A SOTA unsupervised learning method that automatically mines the slot values from unlabeled data by building two neural latent variable models.

\item \textbf{RCAP~\cite{zeng2021automatic}:}
An unsupervised pipeline that works in coarse-to-fine fashion for automatically inferring intents as well as slots labels via sequence labeling, clustering, and Apriori algorithm.

\item \textbf{Inter-Slot~\cite{chen2015jointly}:}
An automatic labeling approach that leverages a random walk inference algorithm
for jointly capturing token-to-token, token-to-slot, and slot-to-slot relationships
by building word-based lexical and slot-based semantic graphs and combining them using dependency grammar.

\item \textbf{Merge-Select~\cite{hudevcek2021discovering}:}
A weak supervision-based approach, which automatically identifies candidate slot values relevant to a given domain using clustering algorithms, then these candidates are used to train a neural network-based slot tagger.

\item \textbf{Coach~\cite{liu2020coach}:}
A zero-shot slot filling model that adapts a coarse-to-fine approach. 
First, it identifies potential slot values irrespective of the slot type and then matches the slot values to appropriate slots using the representation of the slot description.
In this work, we use this model to generate one set of pseudo labels.

\item \textbf{LEONA~\cite{siddique2021linguistically}:}
A zero-shot learning approach, that exploits the embeddings of POS and NER taggers as well as learns a contextual tokens-slot similarity function from seen domains to make predictions about unseen domains.

\item \textbf{GPT-3~\cite{brown2020language}:}
A SOTA model that has shown zero-shot learning capabilities for many NLP tasks. 
In this work, we use this model in the context of few-shot learning and provide five labeled training examples as prompts for each domain (or dataset) and expect the model to make inferences for the given domain. 

\item \textbf{JointBert~\cite{chen2019bert}:}
A supervised learning method that fine-tunes BERT for intent detection and slot filling tasks jointly.
We perform this comparison to understand the performance gap between supervised models that use expensive labeled training data and our proposed zero-label approach. 

\item \textbf{Seq2Seq-DU~\cite{feng2020sequence}:}
a supervised learning approach that formulates the task of dialog understanding as sequence-to-sequence and fine-tunes two BERT encoders, one for utterance and the other for schema descriptions, and jointly models intents and slots.

\end{description}

\section{Results}
\label{results}

\begin{table*}[t!]
\footnotesize
\centering
\caption{Slot F1 score for all the domains in MultiWoZ dataset.}
\vspace{-7pt}
\label{tab:multiwoz-domains}
\begin{tabular}{l|cc|cc|cc|c|cc|c}
\toprule
\multicolumn{1}{c|}{}        & \multicolumn{2}{c|}{\textbf{Unsupervised Learning}}                  & \multicolumn{2}{c|}{\textbf{Weak Supervision}}                             & \multicolumn{2}{c|}{\textbf{Zero-shot Learning}}                & \multicolumn{1}{c|}{\textbf{Few-shot}} & \multicolumn{2}{c|}{\textbf{Supervised Learning}}                        & \multicolumn{1}{c}{\textbf{Self-supervised}} \\
\multicolumn{1}{c|}{\textbf{Domains}} & \multicolumn{1}{c}{\textbf{DSI}} & \multicolumn{1}{c|}{\textbf{RCAP}} & \multicolumn{1}{c}{\textbf{Inter-Slot}} & \multicolumn{1}{c|}{\textbf{Merge-Select}} & \multicolumn{1}{c}{\textbf{Coach}} & \multicolumn{1}{c|}{\textbf{LEONA}} & \multicolumn{1}{c|}{\textbf{GPT-3}}    & \multicolumn{1}{c}{\textbf{JointBERT}} & \multicolumn{1}{c|}{\textbf{Seq2Seq-DU}} & \multicolumn{1}{c}{\textbf{{\ourmodel} (this work)}}            \\ \hline
Hotels                       & 0.6007            & 0.7182                 & 0.4904	& 0.5801             & 0.6481             & 0.7319             & 0.6967   & 0.9292             & \textbf{0.9236}              & \underline{0.9012}          \\
Restaurants                  & 0.5162            & 0.7318                 & 0.4548           & 0.5512             & 0.6137             & 0.7387             & 0.7256   & \textbf{0.9687}             & 0.9475              & \underline{0.8991}          \\
Trains                       & 0.5994            & 0.6637                 & 0.4288	           & 0.5612             & 0.6804             & 0.7926             & 0.5493   & \textbf{0.9524}             & 0.8788              & \underline{0.8741}          \\
Attractions                  & 0.5441            & 0.6446                 & 0.3735          & 0.4305             & 0.3029             & 0.3834             & 0.6057   & 0.9316             & \textbf{0.9880}               & \underline{0.8393}          \\
Taxi                         & 0.5903            & 0.4982                 & 0.2968	& 0.3827                        & 0.1260              & 0.1824             & 0.5175   & \textbf{0.9496}             & 0.7991              & \underline{0.8471}          \\
Others                       & 0.5148            & 0.5990                  & 0.2638	& 0.3132             & 0.1201             & 0.1721             & 0.6689   & \textbf{0.9314}             & 0.9183              & \underline{0.9236}          \\ \hline
Average                      & 0.5609            & 0.6426                 & 0.3847	&0.4698             & 0.4152             & 0.5002             & 0.6273   & \textbf{0.9438}             & 0.9092              & \underline{0.8807}    \\
\bottomrule
\end{tabular}
\vspace{-7pt}
\end{table*}

\begin{table*}[t!]
\footnotesize
\centering
\caption{Slot F1 score for the full datasets: SGD and MultiWoZ.}
\vspace{-7pt}
\label{tab:multiwoz-sgd}
\begin{tabular}{l|cc|ccc|cc|c|cc|c}
\toprule
\multicolumn{1}{c|}{}        & \multicolumn{2}{c|}{\textbf{Unsupervised}}                  & \multicolumn{3}{c|}{\textbf{Weak Supervision}}                             & \multicolumn{2}{c|}{\textbf{Zero-shot Learning}}                & \multicolumn{1}{c|}{\textbf{Few-shot}} & \multicolumn{2}{c|}{\textbf{Supervised Learning}}                        & \multicolumn{1}{c}{\textbf{Self-supervised}} \\
\multicolumn{1}{c|}{\textbf{Datasets}} & \multicolumn{1}{c}{\textbf{DSI}} & \multicolumn{1}{c|}{\textbf{RCAP}} & \multicolumn{1}{c}{\textbf{KG}} &\multicolumn{1}{c}{\textbf{Inter-Slot}} & \multicolumn{1}{c|}{\textbf{Merge-Select}} & \multicolumn{1}{c}{\textbf{Coach}} & \multicolumn{1}{c|}{\textbf{LEONA}} & \multicolumn{1}{c|}{\textbf{GPT-3}}    & \multicolumn{1}{c}{\textbf{JointBERT}} & \multicolumn{1}{c|}{\textbf{Seq2Seq-DU}} & \multicolumn{1}{c}{\textbf{{\ourmodel} (this work)}}            \\ \hline
SGD                          & 0.3352                  & 0.4539        &0.4591                    & 0.3274	& 0.5512                            & 0.1102                    & 0.1621                     & 0.6015                        & \textbf{0.9156}                        & 0.9028                          & \underline{0.8756}                              \\
MultiWoZ                     & 0.4961                  & 0.5937     & 0.4839                      & 0.4483	& 0.5219                            & 0.1903                    & 0.2884                     & 0.6123                        & \textbf{0.9323}                        & 0.8923                          & \underline{0.8423}                              \\ \hline
Average                      & 0.4157                  & 0.5238      & 0.4715                     & 0.3879	& 0.5366                            & 0.1503                    & 0.2253                     & 0.6069                        & \textbf{0.9240}                         & 0.8976                          & \underline{0.8590}     \\
\bottomrule
\end{tabular}
\vspace{-7pt}
\end{table*}


\subsection{Quantitative Analysis}
\label{results-quantitative}
\stitle{Evaluation using a single domain in the dataset.}
Tables~\ref{tab:sgd-domains} and ~\ref{tab:multiwoz-domains} present F1 scores for each domain in \myspecial{SGD} and \myspecial{MultiWoZ} datasets, respectively. 
The best F1 score is shown as bold including the supervised learning methods that were trained using labeled training data for all the domains.
We also highlight the best model by underlining the results excluding the supervised models.
Our proposed framework {\ourmodel} achieves significant performance gains against all the competing models (excluding supervised models) including few-shot GPT-3 on both datasets for every single domain consistently.
Specifically, {\ourmodel} outperforms SOTA models by 20.20\% and 37.05\% on average for all the domains on \myspecial{SGD} and \myspecial{MultiWoZ} datasets, respectively. 
This performance improvement against all the unsupervised learning, weak supervision, zero-shot learning, and few-shot learning approaches can be attributed to the iterative refinement of the noisy labels throughout all the phases of the framework, especially the self-supervised co-training phase.
Moreover, we also notice that our proposed approach which does not use any in-domain labeled data achieves competitive performance to fully supervised models that were trained using labeled training data. 
Specifically, the performance gap is only 2.58 percentage points on average for all the domains in \myspecial{SGD} dataset and 6.31 percentage points for \myspecial{MultiWoZ}.
Interestingly, {\ourmodel} achieves slightly better performance than supervised models for five domains in \myspecial{SGD} dataset.
This incredible performance improvement is due to the outstanding co-training approach of {\ourmodel} that progressively allows the peers to refine the noisy labels. 
Specifically, the zero-shot learning labels carry information from other (seen) domains about how slot values are mentioned across domains, whereas the other set of labels carries the knowledge from KGs and other NLP models. 
The self-supervised co-training phase facilitates combining this information progressively by sharing high-confidence information among peers.
For example, {\ourmodel} achieves a better F1 score than supervised models in the domain \myspecial{Music}. Upon further analysis, we noticed that most of the values for slots \myspecial{song\_name} and \myspecial{artist} are present in the KG, whereas the domain \myspecial{Movies} is quite similar to \myspecial{Music} and the zero-shot model successfully transferred knowledge to {\ourmodel}.

\stitle{Evaluation using the full dataset.}
Table~\ref{tab:multiwoz-sgd} presents the results for the full datasets.
We note that the proposed zero-label framework has, once more, comparable performance to the fully supervised learning methods.
For example, the average F1 score is only 6.5 percentage points lower than the best-performing supervised learning model.
It is important to highlight that this evaluation setup is more challenging because {\ourmodel} automatically generates labels for the whole dataset (i.e., \myspecial{SGD} dataset has 240 slot types) with no human intervention and labeling through heuristic functions or zero-shot learning models is extremely error-prone because of the big number of classes, i.e., 240 in case of \myspecial{SGD} dataset.
To highlight the importance of the gradual noise reduction approach of {\ourmodel}, we also experiment with a KG baseline that uses our approach from phase one to generate pseudo labels and then uses Merge-Select~\cite{hudevcek2021discovering} to further refine the labels.
If we compare our proposed approach to the methods that do not have access to labeled data (or only a few labeled training examples), there is a significant performance gain for {\ourmodel}.
Specifically, {\ourmodel} outperforms unsupervised models by 92.91\% and 41.87\% on \myspecial{SGD} and \myspecial{MultiWoZ} datasets, respectively.
Similarly, other learning paradigms show inferior performance when compared to {\ourmodel}. 
For example, the F1 score of the weak supervision models is 32.44 and 32.04 percentage points lower than our proposed approach for \myspecial{SGD} and \myspecial{MultiWoZ} datasets, respectively.
Last but not least, our performance improvement over SOTA results is statistically significant, with a P-value < 0.01 (i.e., highly significant) for all evaluation settings.

\subsection{Ablation study}
\label{ablation}

\begin{table}[t!]
\centering
\caption{Ablation Study: slot F1 score when certain components are removed from {\ourmodel}}
\vspace{-7pt}
\label{tab:ablation}
\begin{tabular}{lcc}
\toprule
Configuration         & SGD    & MultiWoZ \\
\hline
$\wo$ KG Labels           & 0.8231 & 0.7838   \\
$\wo$ POS / NER Taggers   & 0.7954 & 0.7762   \\
$\wo$ KG, POS/NER Taggers & 0.5767 & 0.5923   \\
$\wo$ Zero-shot Labels    & 0.5091 & 0.4786   \\
$\wo$ Co-training         & 0.5712 & 0.5681   \\
$\wo$ Soft labels         & 0.7145 & 0.6992  \\
\bottomrule
\end{tabular}
\vspace{-7pt}
\end{table}

\stitle{Effect of removing a certain component.} 
Since our proposed framework {\ourmodel} uses several mature components, it is critical to investigate the role of each component.
Table~\ref{tab:ablation} presents the results on both datasets by removing a certain component. 
For example, if we remove KG from phase one and generate automatic labels, the performance slightly declined.
Similarly, removing the pre-trained POS and NER taggers also makes the performance worse.
Moreover, we also note that the effect of removing POS and NER taggers is somewhat greater than KG.   Upon further investigation, we observed that many of the KG entries are eventually automatically labeled in phase two of the framework by the pre-trained BERT.
Moreover, we also look into the result of removing one set of labels entirely from the framework. 
We notice that with only one set of labels, our proposed framework can not utilize phase three effectively and, as a result, shows significantly poor performance.
Moreover, we also note that the effect of removing the zero-shot learning labels is more notable quantitatively as compared to the labels from KG, POS/NER taggers.
The significance of phase three (i.e., self-supervised co-training) is also highlighted by skipping this phase.
In fact, by skipping this phase or not being able to use this phase properly, our proposed framework shows poor performance as compared to other unsupervised or weak supervision learning approaches.
Finally, we also study the effect of high-confidence soft label selection in phase three.
As expected, removing the high-confidence soft label selection mechanism in phase three significantly reduces the performance on both datasets. 
To summarize, {\ourmodel} can still outperform other models that do not use labeled data even after removing these individual components: KG labels, POS/NER taggers, and soft labels.

\stitle{Effect of increasing the out-of-domain data for zero-shot slot filling model.} 
The zero-shot learning models have been shown to improve their performance on new unseen (target) domains, when they are provided with more (out-of-domain) data~\cite{siddique2021linguistically, liu2020coach}. 
We also investigated whether increasing out-of-domain training data for the zero-shot slot filling model improves the performance of our model.
Table~\ref{tab:seen-percentage} presents the F1 score of our proposed framework for both datasets when we vary the percentage of out-of-domain data to train the zero-shot slot filling model.
It can be observed from the results that increasing the out-of-domain training data for the zero-shot learning model improves its prediction accuracy for the target domains, which eventually improves the performance of {\ourmodel}.
Following~\cite{siddique2021linguistically}, we randomly selected domains for out-of-domain training of the zero-shot model and reported average results of five runs.

\begin{table}[t!]
\footnotesize
\centering
\caption{Slot F1 score for different percentages of out-of-domain training data for zero-shot learning model.}
\vspace{-10pt}
\label{tab:seen-percentage}
\begin{tabular}{l|ccc|ccc}
\toprule
Dataset   $\rightarrow$            & \multicolumn{3}{c|}{\textbf{SGD}}  & \multicolumn{3}{c}{\textbf{MultiWoZ}} \\
OOD Seen \% $\rightarrow$ & 25\%   & 50\%   & 75\%   & 25\%     & 50\%    & 75\%    \\ \hline
Coach                 & 0.5888 & 0.6419 & 0.6725 & 0.4408   & 0.4505  & 0.6522  \\
LEONA                 & 0.7181 & 0.7925 & 0.8324 & 0.5248   & 0.5533  & 0.8581  \\ \hline
\textbf{{\ourmodel} (this work)}                  & 0.8592 & 0.9324 & 0.9512 & 0.7937   & 0.8282  & 0.9472 \\
\bottomrule
\end{tabular}
\vspace{-12pt}
\end{table}

\section{Related Work}
\label{related}

\stitle{Supervised Slot Filling.}
Supervised approaches for the slot filling task have been extensively studied.
Recurrent neural networks have been leveraged to learn the temporal interactions of the tokens within a sentence using  Long Short-Term Memory (LSTM) or Gated Recurrent Unit (GRU) networks and detect the spans of the text associated with certain slots~\cite{mesnil2014using,kurata2016leveraging}. 
Moreover, deep learning-based approaches have employed Conditional Random Fields (CRFs) to learn global interactions along with LSTMs and GRUs~\cite{huang2015bidirectional,reimers2017optimal}.
The self-attention technique has also been used for token classification tasks, such as slot filling in the supervised setting~\cite{shen2017disan,tan2017deep}.
Moreover, formulations exist to jointly optimize the tasks of slot filling and intent detection in the supervised setting~\cite{goo2018slot, hakkani2016multi, liu2016attention, zhang2018joint, xu2013convolutional}.
The capsule neural network was trained to jointly model the tasks of slot filling and intent detection in ~\cite{zhang2018joint}, pre-trained BERT was fine-tuned in ~\cite{chen2019bert}, and both tasks were formulated as sequence-to-sequence labeling task in~\cite{feng2020sequence}.
Although supervised approaches have proven effective for the task of slot filling, they require labeled training data for every new domain as well as slot, which prohibits their scalability. In this work, we focus on the open-domain slot filling task and propose a zero-label slot filling framework that can automatically generate labels with no human intervention.

\stitle{Few-shot and Zero-shot Slot Filling.}
To minimize the human labor of annotating data for each new slot, zero-shot learning along with few-shot learning methods have shown promising results for the slot filling task.
Few-shot learning methods rely on a small set of in-domain labeled data to nudge the model for accurate predictions. 
Recently, employing regular expressions~\cite{luo2018marrying}, the prototypical network~\cite{fritzler2019few}, and extensions to CRFs~\cite{hou2020few} have been proposed for the few-shot slot filling task.
Furthermore, several works~\cite{krone2020learning,bhathiya2020meta} have leveraged model agnostic meta-learning to jointly model the slot filling and intent classification tasks.
In a similar line of work, in the zero-shot setting, no in-domain labeled training data is required, which has shown encouraging results to eliminate the need for manual labeling. 
Using LSTM-based models along with the natural language descriptions of the slots has been shown to overcome the challenge of different names of the slot for apparently similar slots (e.g., \myspecial{city} vs \myspecial{destination} in \myspecial{Travel} domain) for the zero-shot adaptation of slot filling task~\cite{bapna2017towards}.
Similarly, attention mechanisms have further improved the performance~\cite{mukherjee2020uncertainty,lee2019zero}.
The concatenation of character and word level embeddings have been employed along with bidirectional LSTMs~\cite{shah2019robust}. 
Moreover, the coarse-to-fine approach in ~\cite{liu2020coach, siddique2021linguistically} was proposed to improve the robustness of the zero-shot slot filling task.
In spite of promising results, there is a huge performance gap between zero-shot learning methods and supervised learning approaches that limit their applicability.
In this work, we employ a zero-shot slot filling model to automatically generate a set of labels in the target domain and further improve the performance of the slot filling task in an open-domain setting.

\stitle{Unsupervised and Weak Supervision Slot Filling.}
The other line of work to eliminate the need for manual labeling is weak supervision and unsupervised learning. 
Unsupervised learning methods~\cite{siddique2021unsupervised} 
have used mining to build latent variable models~\cite{min2020dialogue}.
Moreover, a coarse-to-fine approach using clustering has been explored in~\cite{zeng2021automatic}.
Furthermore, extensions to the DBSCAN~\cite{chatterjee2020intent}, automatic induction of slot values using frame semantics theory~\cite{chen2013unsupervised}, and autoencoders along with hierarchical clustering~\cite{shi2018auto} have been proposed in the unsupervised learning setup.
The weak supervision approaches generally employ a heuristic function~\cite{ren2015clustype,he2017autoentity,fries2017swellshark} to automatically generate labels (i.e., no human intervention) and overcome the challenge of modeling on the noisy labels by employing methods, such as distant-LSTM-CRF~\cite{giannakopoulos2017unsupervised} and dictionaries~\cite{shang2018learning}.
Recently, weak supervision has also been leveraged using clustering algorithms for the automatic selection of candidate values~\cite{hudevcek2021discovering}. Furthermore, random walk inference algorithms were used in~\cite{chen2015jointly} to automatically generate supervision data.
It is important to recall that weak supervision and unsupervised learning models show inferior performance as compared to supervised learning models.
In this work, we employ a GPT-2-based heuristic function to automatically generate one set of labels using KG and other pre-trained NLP models and our proposed framework {\ourmodel} combines this set of labels with the labels generated through the zero-shot model by introducing a self-supervised co-training mechanism.
\section{Conclusion}
\label{conclusion}
We have presented a self-supervised co-training slot filling framework that automatically generates two sets of complementary pseudo labels for co-training two peers in a self-supervised fashion, progressively reduces the amount of noise in the labels by automatically selecting high-confidence soft labels, and improves their prediction power.
In contrast to zero-shot learning and weak supervision approaches that produce inferior quality results as compared to supervised learning approaches, the proposed framework {\ourmodel} achieves competitive performance to supervised slot filling models and shows the promise of open-domain slot filling without any human labeling.
Our extensive evaluations using \myspecial{SGD} and \myspecial{MultiWoZ} datasets that contain dialogs across 20 domains and cover 240 slots, demonstrate that {\ourmodel} outperforms state-of-the-art few-shot GPT-3 by 44.18\% and 40.40\% on average for \myspecial{SGD} and \myspecial{MultiWoZ} datasets, respectively.

\balance

\bibliographystyle{ACM-Reference-Format}
\bibliography{sample-base}


\end{document}